\documentclass[10pt,twocolumn,letterpaper]{article}

\usepackage{cvpr}
\usepackage{times}
\usepackage{epsfig}
\usepackage{graphicx}
\usepackage{amsmath}
\usepackage{amssymb}
\usepackage{wrapfig}
\usepackage{times,color}
\usepackage{cite,footnote,xspace,syntonly,algorithm,algorithmic,bm}

\def \cN {\mathcal{N}}

\def \cM {\mathcal{M}}
\def \cX {\mathcal{X}}
\def \cP {\mathcal{P}}

\def \cD {\mathcal{D}}
\def \cG {\mathcal{G}}
\def \cF {\mathcal{F}}

\def \bPhi {\boldsymbol{\Phi}}
\def \bTheta {\boldsymbol{\Theta}}

\def \by {\mathbf{y}}
\def \bx {\mathbf{x}}

\def \bv {\mathbf{v}}

\def \bI {\mathbf{I}}
\def \bP {\mathbf{P}}

\def \bX {\mathbf{X}}

\usepackage[pagebackref=true,breaklinks=true,letterpaper=true,colorlinks,bookmarks=false]{hyperref}

\cvprfinalcopy 

\def\cvprPaperID{****} 

\ifcvprfinal\fi
\begin{document}

\title{Recurrent Generative Residual Networks for Proximal Learning and Automated Compressive Image Recovery}

\author{Morteza Mardani$^{1}$, Hatef Monajemi$^{2}$, Vardan Papyan$^{2}$, Shreyas Vasanawala$^{3}$, \\ David Donoho$^{2}$, and John Pauly$^{1}$\\
Electrical Engineering$^{1}$, Statistics$^{2}$, and Radiology$^{3}$ Depts., Stanford University\\
{\tt\small morteza,monajemi,papyan,vasanawala,donoho,pauly@stanford.edu}
}

\maketitle

\begin{abstract}
Recovering images from undersampled linear measurements typically leads to an ill-posed linear inverse problem, that asks for proper statistical priors. Building effective priors is however challenged by the low train and test overhead dictated by real-time tasks; and the need for retrieving visually ``plausible'' and physically ``feasible'' images with minimal hallucination. To cope with these challenges, we design a cascaded network architecture that unrolls the proximal gradient iterations by permeating benefits from generative residual networks (ResNet) to modeling the proximal operator. A mixture of pixel-wise and perceptual costs is then deployed to train proximals. The overall architecture resembles back-and-forth projection onto the intersection of feasible and plausible images. Extensive computational experiments are examined for a global task of reconstructing MR images of pediatric patients, and a more local task of superresolving CelebA faces, that are insightful to design efficient architectures. Our observations indicate that for MRI reconstruction, a recurrent ResNet with a single residual block effectively learns the proximal. This simple architecture appears to significantly outperform the alternative deep ResNet architecture by $2$dB SNR, and the conventional compressed-sensing MRI by $4$dB SNR with $100\times$ faster inference. For image superresolution, our preliminary results indicate that modeling the denoising proximal demands deep ResNets.

\end{abstract}

\section{Introduction}
\label{sec:intro}
Linear inverse problems widely appear in image restoration tasks in applications ranging from super-resolving natural images to reconstructing biomedical images. In such applications, one oftentimes encounters a seriously ill-posed recovery task, which necessitates regularization with proper statistical priors. This is however impeded by the following challenges: c1) real-time and interactive tasks afford only a low overhead for inference and training; e.g., imagine MRI visualization for neurosurgery~\cite{clearpoint}, or, real-time superresolution that may need re-training on a cell phone~\cite{romano2017raisr}; c2) the need for recovering plausible images that are consistent with the physical model; this is particularly important for medical diagnosis, which is sensitive to hallucination.

Conventional compressed sensing (CS) relies on sparse coding of images in a proper transform domain via a \textit{universal} $\ell_1$-regularization; see e.g.,~\cite{donoho2006compressed,pualy_mri20017,duarte2009learning}. To automate the time-intensive iterative soft-thresholding algorithm (ISTA) used for sparse coding, \cite{gregor2010learning} proposed learned ISTA (LISTA). Relying on soft-thresholding it trains a simple (single-layer and fully-connected) recurrent network to map measurement to a sparse code. Deep generative networks have also proven tremendously powerful in modeling prior distribution for natural images~\cite{gan-goodfellow2014,johnson2016,lsgan2017,leding_photorealistic_2016,inpainting-yeh-2016,lossfunction_zhao2017}. In particular, residual networks (ResNets) are commonly used due to their stable behavior~\cite{resnet2016} along with pixel-wise and perceptual costs induced by generative adversarial networks (GANs)~\cite{gan-goodfellow2014}. There is a handful of recent attempts to integrate the priors offered by generative nets for inverting linear inverse tasks dealing with \textit{local} image restoration such as super-resolution~\cite{johnson2016,leding_photorealistic_2016}, inpainting~\cite{inpainting-yeh-2016}; and more \textit{global} tasks such as biomedical image reconstruction~\cite{Majumdar_autoencoder_15,sun2016deep,lowdose_ct2017,mardani2017deep,automap_ismrm_2017,partial_fourier_cnn_ismrm_2017,cascade_cnn_mrreocn_ismrm_2017,cs_residual_learning_ismrm_2017}. For instance, deep ResNets are adopted with GANs to superresolve natural images with state-of-the art perceptual quality \cite{johnson2016,leding_photorealistic_2016,Sonderby_2014}.

For biomedical images one typically knows a small fraction of projections onto a certain transform domain (e.g., Fourier, or, Radon) based on physics of the scanner. Variations of deep CNNs are trained to map out aliased MR (AutoMap~\cite{automap_ismrm_2017}) or low-dose CT (RED-CNN~\cite{lowdose_ct2017}) images to the gold-standard ones retrieved by iterative CS. They offer rapid reconstruction at the expense of high training overhead. There is however no systematic mechanism to assure fidelity to the underlying physical model, which can possibly hallucinate images. Another line of work pertains to developing effective priors to incorporate in an iterative algorithm which can outperform the conventional sparsity priors for CS; see e.g.,~\cite{Bora_dimakis_2017,hand2017global,sun2016deep,adler2017learned,adler2017learning}. For instance, \cite{Bora_dimakis_2017,hand2017global} uses the low-dimensional code offered by a pre-trained generative decoder to achieve higher SNR than CS. These schemes attain a reasonably high SNR, but need several iterations for convergence that hinders real-time imaging.

Toward rapid, feasible, and plausible image recovery for ill-posed linear inverse tasks, this paper proposes a novel approach to automate imaging. Inspired by proximal gradient iterations, a recurrent ResNet architecture is designed to learn the proximal(s) from the data. Image prior information is learned through the proximal, that is modeled as a generator (G) network, consisting a few residual blocks (RB), trained with mixture of pixel-wise and perceptual costs via GANs to recover plausible images. The overall architecture implements multiple back-and-forth (approximate) projections onto the subspace of data-consistent images - dictated by the physical model - and the manifold of plausible images. The number of projections and similarly the size of each G network is desired to be small to reduce the training and inference overhead for real-time and interactive image recovery tasks. To study this effect, we perform experiments for reconstructing pediatric MR images, and super-resolving natural images from the CelebA face dataset. The former is a global task with subsampled measurements in the frequency domain, so-termed $k$-space, while the latter is a rather local task, with a pixelated low-resolution image.

Our observations indicate that for MRI reconstruction, one better repeat a small ResNet (with a single RB) multiple times, instead of training a deep network, as commonly used e.g., in \cite{automap_ismrm_2017,partial_fourier_cnn_ismrm_2017,cascade_cnn_mrreocn_ismrm_2017,cs_residual_learning_ismrm_2017}. The recurrent architecture not only improves upon the deep schemes by about $2$dB SNR, but also incurs much less training overhead. This simple architecture also significantly outperforms the conventional CS-Wavelet/CS-TV schemes; $4$dB SNR gain in $100$ times shorter time. For the local $4$x superresolution task however it turns out that one needs to train a deep ResNets to learn the denoising proximals, and the preliminary results do not show any major advantage in using recurrent schemes. Before closing this section, it is worth to re-iterate the main contributions of this paper as follows:

\begin{itemize}

\item Learning proximals using ResNet generators trained with pixel-wise and GAN-based perceptual costs

\item Extensive experiments for MRI reconstruction and face superresolution gives insight to design proper network architectures for various recovery tasks

\end{itemize}

The rest of this paper is organized as follows. Section~\ref{sec:problem_state} states the problem. Proximal learning based on recurrent GANs is discussed in Section~\ref{sec:proximal_learn}. Evaluations for pediatric MR image reconstruction, and natural image superresolution are reported in Section~\ref{sec:eval}, while the conclusions are drawn in Section~\ref{sec:conc}.



\section{Preliminaries and problem statement}
\label{sec:problem_state}
Consider an ill-posed linear system $\by=\bPhi \bx + \bv$ with $\bPhi \in \mathbb{C}^{M \times N}$ where $M \ll N$, and $\bv$ captures the noise and unmodeled dynamics. Suppose the unknown and (complex-valued) image $\bx$ lies in a {\it low-dimensional} manifold, say $\cM$. No information is known about the manifold besides the training samples $\cX:=\{\bx_k\}_{k=1}^K$ drawn from it, and the corresponding (possibly) noisy observations $\mathcal{Y}:=\{\by_k\}_{k=1}^K$. Given a new undersampled observation $\by$, the goal is to \textit{quickly} recover a plausible $\bx$.

The stated problem covers a wide range of image restoration and reconstruction tasks. For instance, in medical image reconstruction $\bPhi$ describes a projection driven by physics of the acquisition system (e.g., Fourier transform for MRI scanner). For image superresolution it is the downsampling operator that averages out nonoverlapping image regions to arrive at a low-resolution one. Given the image prior distribution, one typically forms a maximum-likelihood estimator formulated as a regularized least-squares (LS) program
\begin{align}
{\rm (P1)}  \quad \quad \min_{\bx}~~\big{\|}\by - \bPhi \bx\big{\|}^2 + \psi(\bx;\bTheta)   
\end{align}
with the regularizer $\psi(\cdot)$ parameterized by $\bTheta$ that incorporates the image prior.

In order to solve (P1) one can adopt a variation of proximal gradient algorithm~\cite{parikh2014proximal} with a proximal operator $\cP_{\psi}\{\cdot\}$ that is obtained based on $\psi(\cdot)$~\cite{parikh2014proximal}. Starting from $\bx[0]=\mathbf{0}$, and adopting a small step size $\alpha$ the overall iterative procedure is expressed as
\begin{align}
\bx[k+1] &= \cP_{\psi} \Big\{ \bx[k] + \alpha \bPhi^{\mathsf{H}} (\by -  \bPhi \bx[k])  \Big\}   \nonumber\\ 
&=  \cP_{\psi} \Big\{ \alpha\bPhi^{\mathsf{H}} \by +  \underbrace{(\bI -  \alpha \bPhi^{\mathsf{H}}\bPhi )}_{:=\mathbf{P}_{\cN}}\bx[k]  \Big\}   \label{eq:gradient_proj}
\end{align}

For convex proximals the fixed point of \ref{eq:gradient_proj} coincides with the global optimum for (P1)~\cite{parikh2014proximal}. For some simple prior distributions, the proximal operation is convex and tractable in closed-form. One popular example of such a proximal pertains to $\ell_1$-norm regularization for sparse coding, where the proximal operator gives rise to soft-thresholding and shrinkage in a certain domain such as Wavelet, or, total-variation (TV). The associated iterations goes by ISTA, which are then improved to FISTA iterations with accelerated convergence~\cite{beck2009fast}.

As argued earlier FISTA is a universal and thus naive regularization that does not take into account the image complications and perceptual quality. In addition, for moderate and high resolution images it demands many iterations for convergence that can seriously imped real-time recovery. The next sections aim to fix these caveats by learning proximals from historical images using generative neural networks.


\section{Proximal learning}
\label{sec:proximal_learn}
Motivated by the proximal gradient iterations in \eqref{eq:gradient_proj}, to design efficient network architectures that automatically invert linear tasks, we need to first address the following important questions:
\begin{itemize}
\item How can one ensure the network does not hallucinate images, and retrieves plausible images that are physically feasible?

\item How can one ensure rapid inference and affordable training for real-time and interactive image recovery tasks?
\end{itemize}
%





\subsection{Recurrent network architecture}
The recursion in \eqref{eq:gradient_proj} can be envisioned as a feedback loop in Fig. 1, which takes an initial image estimate $\tilde{\bx}$ that is subsequently projected onto the manifold $\cM$ through the proximal operator to return $\hat{\bx}$. The proximal operator is modeled via a generative neural networks as will be elaborated in the next section. Projection onto $\cM$ is supposed to remove artifacts to some extent, and result in a more visually appealing image. To close the loop, $\hat{\bx}$ then passes through the filter $\bP_{\cN}:=\bI - \alpha \bPhi^{\mathsf{H}}\bPhi$ and is added up to the input  $\tilde{\bx}$. The feedback filter $\bP_{\cN}$ resembles projection onto the nullspace of the measurement operator. Notice, $\bP_{\cN}$ only needs tunning the step size $\alpha$ that can be easily learned from the data. Implementing the feedback loop however demands (possibly) infinitely many iterations.

To bypass this hurdle, inspired by recurrent neural networks (RNNs) we unroll the loop and repeat multiple, say $K$, copies of the proximal network as depicted in Fig. 1 (bottom). Each proximal network is accompanied with an (approximate) data consistency projection that refines $\check{\bx}$ to be consistent with the observations by simply moving along the descent direction of the data fidelity cost, namely $\|\by-\bPhi\bx\|^2$. Assuming exact data consistency projection the unrolled network learns the projection onto the intersection of \textit{physically feasible} and \textit{visually plausible} images. In general one can consider the cascaded architecture in Fig. 1 with independent weights $\{\bTheta_k\}_{k=1}^K$ per copies, but we are more interested in sharing the weights, namely $\bTheta_1=\ldots=\bTheta_K$, which needs less training variables, and the back-propagation can easily accommodate gradient calculations~\cite{Goodfellow-et-al-2016}.

In essence, (multiple) back-and-forth projections can ensure data fidelity to a good extent. This is in contrast with the existing deep architectures for automated medical image reconstruction (e.g.,~\cite{Majumdar_autoencoder_15,automap_ismrm_2017,lowdose_ct2017}) with no consideration for data fidelity, which may hallucinate images and mislead the diagnosis. It is also worth mentioning that the Amortised-MAP based deep-GAN scheme in~\cite{Sonderby_2014} uses an affine projection layer that improves GAN's stability and suerresolution quality. However, one naturally need multiple projections to assure data consistency. The number of copies however cannot be large, or, alternatively the proximal networks need to be small, for real-time inference tasks. 


\begin{figure*}[t]
	\centering
  \includegraphics[scale=0.75]{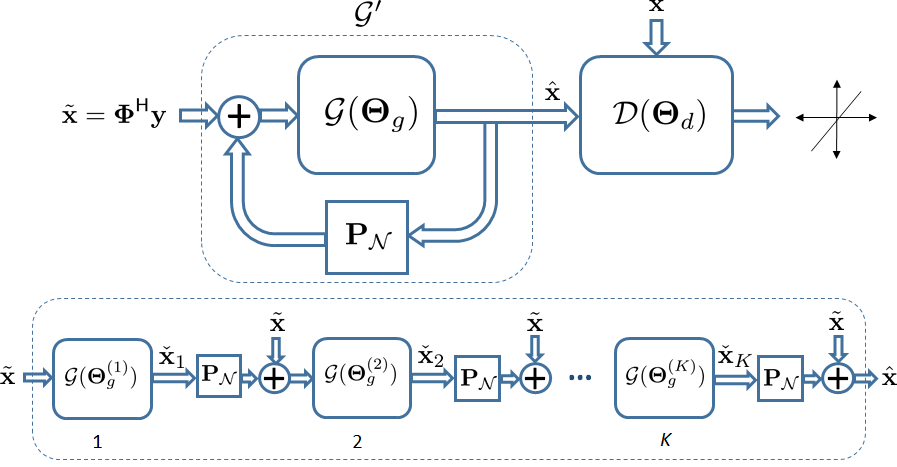}
	\caption{Recurrent GAN. (Top) the closed-loop circuit diagram, and (right) unrolled computational graph with $K$ copies. }
	\label{fig:fig_net}
\end{figure*}

\subsection{Mixture of pixel-wise and perceptual costs}
\label{subsec:GAN_manifold}
To learn proximals as projections onto manifold of visually plausible images we adopt GANs~\cite{gan-goodfellow2014}. Conventional generative models such as variational auto-encoders~\cite{leding_photorealistic_2016} rely on pixel-wise costs that offer high pick signal-to-noise ratios but often produce overly-smooth images with poor perceptual quality. GANs however train a perceptual loss from the training data. Standard GANs consist of a tandem structure of generator (G) and discriminator (D) networks~\cite{gan-goodfellow2014}. 


Training GANs amounts to playing a game with conflicting objectives between the adversary G and the discriminator. D network aims to score one the training ground-truth images drawn from the data distribution, and zero the (fake) outputs of G. Apparently, D cannot perfectly separate real and fake images as G tries to generate fake images that fools G. Various strategies have been devised to reach the game's equilibrium. They mostly differ in evaluating the loss incurred by G and D~\cite{gan-goodfellow2014}, \cite{lsgan2017}. The conventional GAN~\cite{gan-goodfellow2014} uses a sigmoid cross-entropy for D's loss, which suffers from vanishing gradients. It leads to unstable training that causes mode collapse. In addition, for the generated images classified confidently as real (with a large decision variable), no cost is incurred. Hence, it tends to pull samples away from the decision boundary, that introduces non-realistic images~\cite{lsgan2017}. This particularly can hallucinate medical images, and as a result mislead medical diagnosis. To alleviate this issue, we adopt least-square GAN (LSGAN) that penalizes the classification mistake with a LS cost that pulls the generated samples towards the decision boundary.

One issue with GAN however is that it may over-emphasize high frequencies at the expense of deteriorating the image main structure. To avoid this issue, along with LSGAN perceptual loss we use a pixel-wise $\ell_1$/$\ell_2$ cost, which perform well in maintaining the structure, and discarding the low-intensity noise~\cite{lossfunction_zhao2017}. Training G with the mixture cost is thus expected to reveal fine texture details while discarding noise. Let us collect all the network parameters in $\bTheta:=(\bTheta_g, \bTheta_d)$. The overall procedure then aims to jointly minimize the discriminator cost
\begin{align}
&{\rm(P1.1)} \quad \min_{\bTheta}~ \mathbb{E}_{\bx} \Big[\Big(1-\cD(\bx;\bTheta_d)\Big)^2\Big]   +  \mathbb{E}_{\by} \Big[\Big(\cD(\hat{\bx};\bTheta_d)\Big)^2\Big]  \nonumber
\end{align}
and the generator cost
\vspace{-3mm}
\begin{align}
&{\rm(P1.2)}~ \min_{\bTheta} \mathbb{E}_{\by}\Big[ \sum_{k=1}^K \big\|\by - \bPhi \check{\bx}_k \big\|^2 \Big] \hspace{-1mm}+\hspace{-1mm} \lambda \mathbb{E}_{\by} \Big[\Big(1-\cD \big(\hat{\bx};\bTheta_d\big)\Big)^2\Big] \nonumber \\& \hspace{3cm} + \eta \mathbb{E}_{\bx,\by} \Big[ \big\| \mathbf{x} - \hat{\bx}\big\|_{1,2} \Big]
\end{align}
where $\|\bx\|_{1,2}:=\gamma \|\bx\|_{1} + (1-\gamma) \|\bx\|_{2}$ for some $0 \leq \gamma \leq 1$. The LS data fidelity term in (P1.2) is a soft version of the affine projection in the network architecture of Fig. 1. Parameter $\lambda$ is also tuned based on the measurement noise level and the expected pixel-wise fidelity.

\section{Experiments}
\label{sec:eval}
Performance of the novel recurrent GANCS scheme is assessed in reconstructing pediatric MR images and super-resolving natural images. The former introduces aliasing artifacts that globally impact the entire image pixels, while in the latter the pixelation occurs locally. While the focus is mostly placed on MRI, preliminary results are also reported for image super-resolution to shed some light on challenges associated with proximal learning. In particular, we aim to address the following intriguing questions:

\vspace{2mm}


\noindent\textbf{Q1.}~What is the proper number of copies, and generator size to learn the proximal?

\noindent\textbf{Q2.}~What is the trade-off between PSNR/SSIM and inference/training complexity?

\noindent\textbf{Q3.}~How is the performance compared with the conventional sparse coding?

\noindent\textbf{Q4.}~How does the performance change if we train with independent weights per copies, and what is the interpretation for output of different copies?

\subsection{Residual generator networks and training}
\label{subsec:training}
\begin{figure*}[t]
	\centering
	\hspace{-0.0cm}\includegraphics[scale=0.95]{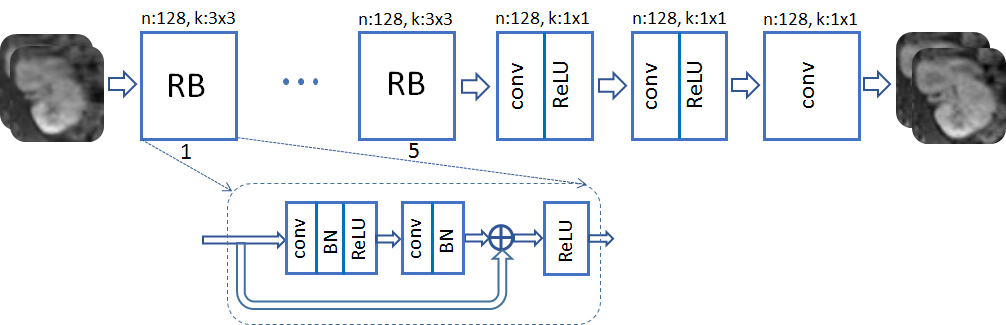}
	\caption{Generator ResNet architecture with RBs, $n$ and $k$ refer to number of feature maps and filter size, respectively. }
    
	\label{fig:fig_gen_net}
\end{figure*}

To address the above questions, for the generator networks we adopt a ResNet with a variable number of residual blocks (RB). Each RB consists of two convolutional layers with $3 \times 3$ kernels and a fixed number of $128$ feature maps, respectively, that are followed by batch normalization (BN) and ReLU activation. It is then followed by three simple convolutional layers with $1 \times 1$ kernels, where the first two layers undergo ReLU activation and the last layer has sigmoid activation to return the output; see Fig.~\ref{fig:fig_gen_net}. Notice that for all generators $\{\cG(\bTheta_k)\}_{k=1}^K$ a similar ResNet architecture is used.

The D network is composed of eight convolutional layers. In all the layers except the last one, the convolution is followed by BN and ReLU activation. No pooling is used. For the first four layers, number of feature maps is doubled from $8$ to $64$, while at the same time convolution with stride $2$ is used to reduce the image resolution. Kernel size $3 \times 3$ is adopted for the first five layers, while the last two layers use kernel size $1 \times 1$. In the last layer, the convolution output is averaged out to form the decision variable for LS binary classification, where no soft-max is used.

Adam optimizer is used with the momentum parameter $\beta=0.9$, mini-batch size $L_b=2$, and learning rate $\mu=10^{-5}$. Training is performed with TensorFlow interface on a NVIDIA Titan X Pascal GPU with 12GB RAM.



\begin{figure*}[t]
	\centering
	\hspace{-0.0cm}\includegraphics[scale=0.925]{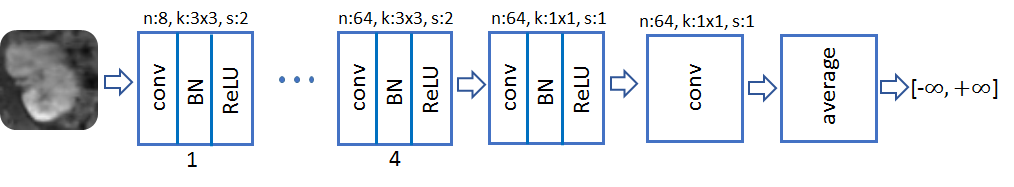}
	\caption{Discriminator multilayer CNN architecture with the input magnitude image, where $n$, $k$, and $s$ refer to number of feature maps, filter size, and stride size, respectively. }
	\label{fig:fig_disc_net}
\end{figure*}

\subsection{MRI reconstruction and artifact suppression}
\label{subsec:artifact_supression}
\begin{figure}[t]
	\centering
	\hspace{-0.0cm}\includegraphics[scale=0.525]{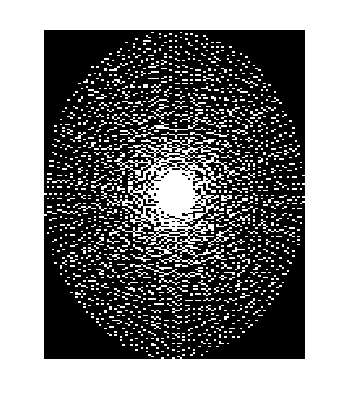}
	\caption{$k$-space sampling trajectory with $5$-fold undersampling based on variable density distribution with radial view ordering. }
	\label{fig:sampling_trajectory}
\end{figure}

Performance of the novel recurrent scheme is assessed in removing aliasing artifacts from MR images. In essence, the scanner acquires Fourier coefficients ($k$-space data) of the underlying image across various coils. A single-coil MR acquisition model is considered where for $n$-th patient the acquired $k$-space data admits 
\begin{align}
y_{i,j}^{(n)} = [\cF(\bX_n)]_{i,j} + v_{i,j}^{(n)},~~(i,j) \in \Omega
\end{align}
Here, $\cF$ refers to the 2D Fourier transform, and the set $\Omega$ indexes the sampled Fourier coefficients. As it is conventionally performed with CS MRI, we select $\Omega$ based on a variable density sampling with radial view ordering that is more likely to pick low frequency components from the center of $k$-space~\cite{pualy_mri20017}. Only $20\%$ of Fourier coefficients are collected. The sampling mask is shown in Fig.~\ref{fig:sampling_trajectory}.

\noindent{\textbf{Dataset}.}~Contrast-enhanced T1-weighted abdominal image volumes are acquired for $350$ pediatric patients. Each 3D volume includes $151$ axial slices of size $200 \times 100$ pixels. $300$ patients ($45,300$ slices) are considered for training, and $50$ patients ($7,550$ slices) for test. All in vivo scans were acquired on a 3T MRI scanner (GE MR750) with voxel resolution $1.07 \times 1.12 \times 2.4$ mm. The input and output are complex-valued images of the same size and each include two channels for real and imaginary components. The input image $\tilde{\bx}$ is simply generated using inverse 2D FT of the $k$-space data where the missing ones are filled with zero (ZF), and is severely contaminated with artifacts. In fact one may ponder whether the gold-standard, i.e., the fully-sampled raw $k$-space data, is available. The answer is affirmative. We average out the acquired time-resolved contrast-enhanced $k$-space data over time to end up with fully-sampled data. Then, we synthesize the undersampled data by randomly selecting only $20\%$ of the $k$-space pixels based on the sampling mask in Fig.~\ref{fig:sampling_trajectory}.




\subsubsection{Performance for various number/size of copies}
\label{subsubsec:mri_trade_off}
In order to assess the impact of network wiring on the image recovery performance, the cascaded network is trained for a variable number of ResNet copies with variable number of RBs. $10$k slices from the train dataset set are randomly picked for training, and $1,280$ slices from the test dataset for test.

\begin{figure*}[t]
    \centering
	\includegraphics[scale=0.875]{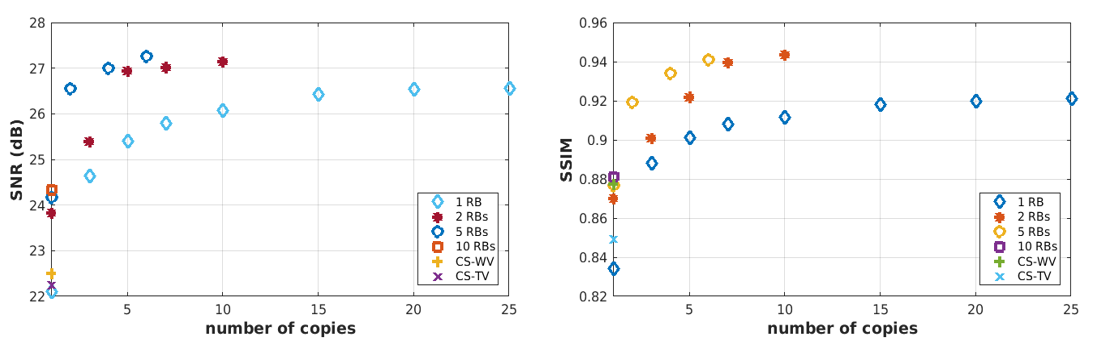} 
	\caption{Average SNR and SSIM versus the number of copies when the weights are shared among the copies. }
	\label{fig:fig_snr_ssim_copy_mse}
\end{figure*}

\noindent\textbf{Shared training weights.}~The copies are first assumed to be identical with shared weights that results in a recurrent network. Training is performed for various combinations of GAN and pixel-wise costs. We report most of the results with only $\ell_2$ cost alone in (P1), i.e., $\gamma=0,\eta=1,\lambda=0$, where no D network is trained as it is easier to train and evaluate the performance quantitatively for several network architectures, which is an important purpose of this work. We report more evaluations with GAN perceptual cost later. Fig.~\ref{fig:fig_snr_ssim_copy_mse} depicts the SNR and structural similarity index metric (SSIM)~\cite{wang2004image} versus the number of copies, when each copy comprises $1/2/5/10$ RBs. It is observed that increasing the number of copies significantly improves the SNR and SSIM, but lead to a longer inference and training time. In particular, using three copies instead of one achieves more than $2$dB SNR gain for $1$ RB, and more than $3$dB for $2$ RBs. It is also interestingly observed that when using a single copy, adding more than $5$ RBs to make a deeper network does not improve anymore; look at the SNR=$24.33$ for $10$ RBs, and SNR=$24.15$ for $5$ RBs. Notice also that a single RB is not also expressive enough to learn the MR image denoising proximal, and as a result repeating it several times, the SNR does not seem to exceed $27$dB. Using $2$ RBs however turns out to learn the proximal, and perform as good as using $5$ RBs. Similar observations are made for SSIM.


\begin{figure*}[t]
\centering
	\includegraphics[scale=0.9]{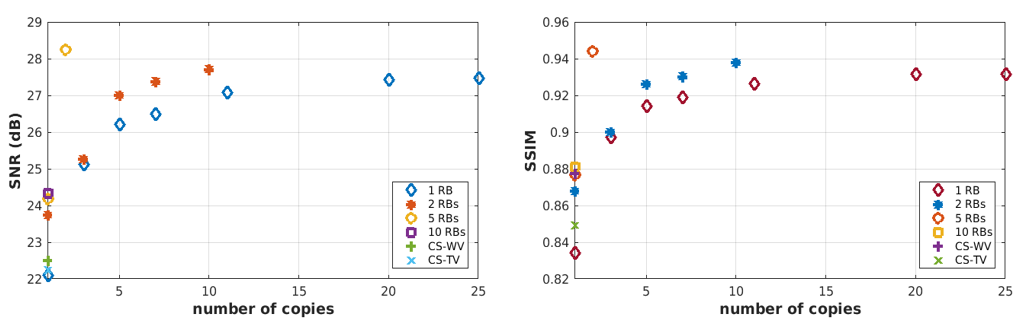} 
	\caption{Average SNR and SSIM versus the number of copies when the weights across copies are allowed to change independently. }
	\label{fig:fig_snr_ssim_copy_mse_different}
\end{figure*}

\noindent\textbf{Independent weights.}~We also consider a scenario where one allows weights varying across different copies. A similar ResNet architecture is used for all copies, which multiplies the variable count for training by the number of copies. As seen in Fig.~\ref{fig:fig_snr_ssim_copy_mse_different}, adopting a single RB per copy and repeating it for $15-20$ copies seems to be a suitable choice that achieves up to $27.4$dB SNR. Apparently, a single RB and $10$ copies performs as good as $4-5$ RBs with $5$ copies in terms of SNR and SSIM. Comparing with the shared weight scenario, for $10$ copies with a single RB, using independent weights improves the SNR by almost $1$dB. Notice that when each copy includes more than $5$ RBs, our GPU resources become exhausted for more than $5$ copies, and thus the rest of points are not shown on the plot. Further evaluations with more efficient implementation and stronger GPU resources is deferred for our future research. 



\noindent\textbf{Training and inference time.}~Inference time for both shared and independent weights is the same, and proportional to the number of copies. Feed-forwarding each image through a copy with one RB takes  $4$ msec when fully using the GPU. The training variable count is also proportional to the number of copies when the weights are allowed to change per different copies. It is hard to precisely evaluate the training and inference time under fair conditions as it strongly depends on the implementation and the allocated memory and processing power per run. As an estimate for the inference time we average it out over a few runs on the GPU as listed in Table~\ref{tab:table_train_time}. It is empirically observed that with shared weights, e.g., $10$ copies with $1$ RB the training converges in $2-3$ hours, but a deep single copy ResNet with $10$ RBs takes around $10-12$ hours to converge.



\begin{table*}[t]
	\caption{Performance trade-off for various architectures with shared and independent weights. }
	\vspace{0.5mm}
	\label{tab:table_train_time}
	\begin{center}
    \begin{tabular}{|c|c|c|c|c|c|c|c|c|}
			\hline
            copies & RBs   & inference time (sec) & SNR (dB), independent & SSIM, independent & SNR (dB), shared & SSIM, shared\\
			\hline\hline
            $10$ & $1$   & $0.04$ & $27.03$  & $0.923$ & $26.07$  & $0.9117$ \\
			\hline
            $5$ & $2$  & $0.10$ & $27.01$  & $0.9258$ & $26.94$  & $0.9221$\\
			\hline
			$2$ & $5$  & $0.12$ & $28.14$  & $0.944$ & $26.55$  & $0.9194$\\  
            \hline
			$1$ & $10$  & $0.0522$ & $24.33$  & $0.8810$ & $24.33$  & $0.8810$\\
			\hline
			CS-TV & n/a & $1.30$ & $22.20$  & $0.82$ & $22.20$  & $0.82$\\
			\hline
            CS-WV & n/a &  $1.16$ & $22.51$  & $0.86$ & $22.51$  & $0.86$\\
			\hline
		\end{tabular}
	\end{center}
\end{table*}

\begin{figure*}[t]
    \centering
	\includegraphics[scale=1.0]{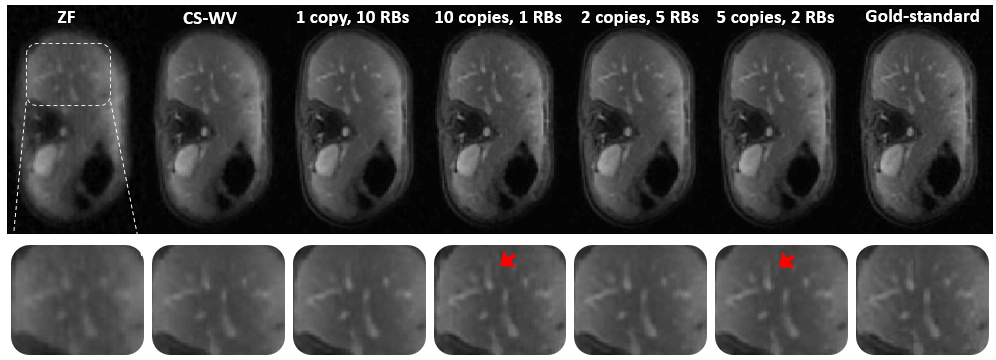} 
	\caption{A representative axial abdominal slice for a test patient reconstructed by zero-filling (1st column); CS-WV (2nd column); and RGANCS with $1$ copy and $10$ RBs (3rd column), $10$ identical copies and $1$ RBs (4th column), $2$ identical copies and $5$ RBs (5th column), $5$ identical copies and $2$ RBs (6th column); and the gold-standard (7th column).}
	\label{fig:fig_rep_slices_diff_netarchs_vs_cs_identicalcopies}
\end{figure*}

\subsubsection{Comparison with sparse coding}
\label{subsubsec:comparison_cs}
To compare with the conventional CS schemes, CS-WV and CS-TV are adopted and tunned for the best SNR performance using BART~\cite{bart2016} that runs $300$ iterations of FISTA along with $100$ iterations of conjugate gradient descent to reach convergence. Quantitative results are listed under Table~\ref{tab:table_train_time}, where it is evident that the recurrent scheme with shared weights significantly outperforms CS with more than $4$dB SNR gain that leads to sharper images with finer texture details as seen in Fig.~\ref{fig:fig_rep_slices_diff_netarchs_vs_cs_identicalcopies}. As a representative example Fig.~\ref{fig:fig_rep_slices_diff_netarchs_vs_cs_identicalcopies} depicts the reconstructed abdominal slice of a test patient. CS-WV retrieves a blurry image that misses out the sharp details of the liver vessels. A deep ResNet with one copy and $10$ RBs captures a cleaner image, but still smoothens out fine texture details such as vessels. However, when using $10$ simple copies with a single RB, more details are seen about the liver vessels, and the texture appears to be more realistic. Similarly, using $5$ copies each containing $2$ RBs retrieves finer details than $2$ relatively large copies with $5$ RBs.

This observation indicates that the proximal for denoising MR images is well represented by a small number $1-2$ RBs. The important message however is that multiple back-and-forth iterations are needed to recover a plausible MR image that is physically feasible. considering the training and inference overhead as well as the quality of reconstructed image in Fig.~\ref{fig:fig_rep_slices_diff_netarchs_vs_cs_identicalcopies}, the architecture with $10$ copies and $1$ RB seems promising to implement in clinical scanners.

\subsubsection{LSGAN for sharp MR images}
\label{subsubsec:lsgan_sharp}
We train the GAN scheme with the generator cost (P1.2) that relies $90\%$ on the pixel-wise $\ell_2$ cost, and $10\%$ on the LSGAN cost. Recurrent LSGAN is trained with shared weights per copies. To avoid mode collapse, we begin the training with the pixel-wise $\ell_2$ cost and gradually increase the GAN loss weight to reach $10\%$ after around $10^3$ mini-batches. This helps the GAN output to be consistent with the images, and thus the generator initial distribution overlapping with the true image distribution. We found this trick very useful in stably training GANs especially for the proposed recurrent architecture. All the network architectures discussed next are seen to converge after a few dozens epochs over the training data.

Fig.\ref{fig:fig_rep_slices_diff_netarchs_vs_cs_identicalcopies} compares the retrieved images by various recurrent GAN architectures with the input ZF image as well as the gold-standard one that is fully-sampled. Abdominal slices shown for two representative axial slices including liver and kidneys confirm again that RGANCS scheme with $10$ copies and $1$ RB performs the best in terms of perceptual quality. Even though SNR and SSIM are not proper metrics to assess the perceptual quality, for the sake of completeness we report them in Table~\ref{tab:table_train_time}. This also corroborates even when using perceptual loss for training, recurrent scheme can significantly improve SNR/SSIM relative to a single deep network ($1$ copy, $10$ RBs) as commonly adopted for image restoration tasks in the literature. The RGANCS images are sharper than the CS-wavelet scheme, even though CS achieves a higher SNR/SSIM. Choosing a smaller weight $\lambda$, or, a larger $\eta$, RGANCS can even improve the SNR/SSIM as it was seen in Table 1 of the paper. Further tunning of $\lambda$ and $\eta$ for the best performance needs expert opinion of radiologists about the diagnostic quality of the resulting images and is the subject of our ongoing research.

\begin{figure*}[t]
    \centering
	\includegraphics[scale=1.25]{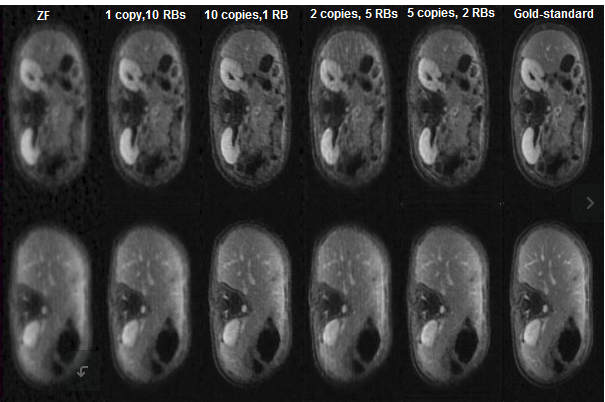} 
	\caption{Representative axial abdominal slices for a test patient reconstructed by zero-filling (1st column); and  the recurrent GANCS (RGANCS) with $1$ copy and $10$ RBs (2nd column), $10$ identical copies and $1$ RBs (3rd column), $2$ identical copies and $5$ RBs (4th column), $5$ identical copies and $2$ RBs (5th column); and the gold-standard (6th column). For RGANCS we used $\eta=0.9$ and $\lambda=0.1$.}
	\label{fig:fig_rep_slices_diff_netarchs_vs_cs_identicalcopies}
\end{figure*}

\begin{table}[t]
	\caption{SNR and SSIM performance for various architectures with shared weights when $\eta=0.9$ and $\lambda=0.1$. }
	\vspace{0.5mm}
	\label{tab:table_train_time}
	\begin{center}
    \begin{tabular}{|c|c|c|c|c|c|c|c|c|}
			\hline
            copies & RBs   & SNR (dB) & SSIM\\
			\hline\hline
            $10$ & $1$   & $22.04$ & $0.835$  \\
			\hline
            $5$ & $2$  & $20.16$ & $0.794$  \\
            \hline
            $2$ & $5$  & $19.24$ & $0.7496$  \\  
            \hline
			$1$ & $10$  & $18.48$ & $0.71$  \\
			\hline
		\end{tabular}
	\end{center}
\end{table}



\subsection{Single image super-resolution}
\label{subsec:image_superresolution}
More evaluations are performed for super-resolving natural images. In essence, super-resolution can be seen as a linear inverse task, where one has only access to a low-resolution image $\by=\phi * \bx + \bv$ obtained after downsampling with a convolution kernel $\phi$. We adopt a $4 \times 4$ constant kernel with stride $4$ that averages out the image pixel intensities over $4\times 4$ non-overlapping regions. Image super-resolution is a challenging ill-posed problem, and has thus been the subject of intensive research over the last decade; see e.g.,~\cite{bruna2015super,Sonderby_2014,johnson2016,romano2017raisr} and the references therein. \cite{Sonderby_2014} leverages deep convolutional GANs (DCGANs) accompanied with an affine projection layer to find a better solution as measured by SNR and SSIM. \cite{leding_photorealistic_2016} also deploys a deep ResNet ($16$ RBs with $64$ feature maps) along with GAN perceptual cost to retrieve photo-realistic images. Our goal is not to create better looking images than the state-of-the-art, but to study proximal learning for this application that gives insights about possibly simpler network architectures for real-time tasks, and can interpret the proximal behavior in terms of revealing the details.




\noindent\textbf{CelebA dataset.}~Adopting celebFaces Attributes Dataset (CelebA)~\cite{liu2015faceattributes}, for training and test we use $10$k and $1,280$ images, respectively. Each ground-truth face image has $128 \times 128$ pixels that is down-sampled to a $32 \times 32$ low-resolution image.



\noindent\textbf{Training.}~Our TensorFlow implementation uses a 2D conv with stride $4$ for downsampling, and transpose conv with stride $4$ for upsampling. Note, transpose convolution does not perform deconvolution, and a single conv transpose can be quite suboptimum. We approximate the deconvolution pseudo-inverse with a few ($5$) gradient-descent iterations with a small step size ($0.1$). The deconvolution then involves conv and transpose conv, and its gradient turns out to below up abruptly and generates NaNs during gradient backpropagation. We thus use gradient clipping for the G network that fixes the issue. The network is fed with pixelated $128 \times 128$ images with three RGB channels obtained by an approximate deconvolution. The same network architecture as for MRI is adopted. 



\begin{figure*}[t]
\centering
	\includegraphics[scale=0.9]{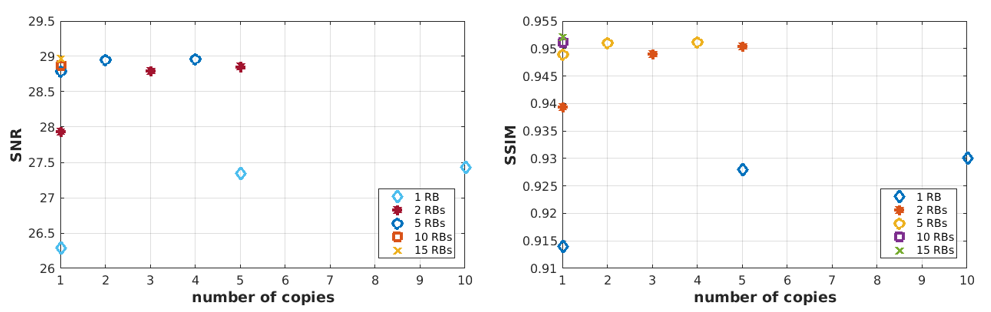} 
	\caption{Average SNR and SSIM for super-resolving CelebA face images for various number of copies when the weights are independent. }
	\label{fig:fig_snr_copy_mse_superresolution}
\end{figure*}

\begin{figure*}[t]
    \centering
	\hspace{0cm}\includegraphics[scale=1.15]{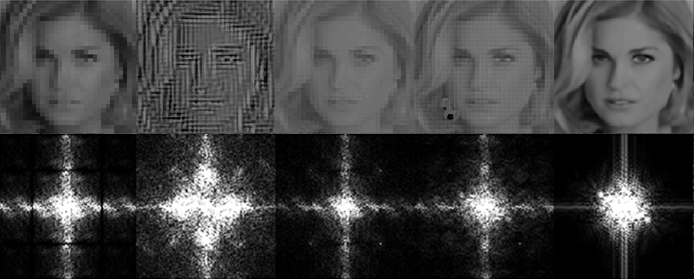} 
	\caption{From left to right: low-resolution input $\tilde{\bx}$, generator outputs $\check{\bx}_1,\check{\bx}_2,\check{\bx}_3,\hat{\bx}$, respectively, for a representative test face image. All images are displayed in gray scale. First row shows the output images with the corresponding frequency domain in the second row, when trained using $4$ independent ResNet copies with $5$ RBs.}
	\label{fig:fig_4outputs_5RBs_batc12_superresolution}
\end{figure*}

\vspace{-2mm}

\subsubsection{Local versus global recovery tasks}
\label{subsubsec:local_vs_global}
For the superresolution task when sharing the weights no interesting pattern is observed for ResNets of size $1-7$ RBs, which indicates modeling the proximal needs larger networks. For the scenario with varying weights Fig.~\ref{fig:fig_snr_copy_mse_superresolution} plots PSNR and SSIM for various architectures. Using $5$ copies with $2$ RBs seems to perform as good as a deep ResNet with $15$ RBs adopted in~\cite{leding_photorealistic_2016}. However, increasing the number of copies and RBs does not offer any clear advantages. This is in contrast with MRI reconstruction where a recurrent single RB could significantly outperform the deep architectures with up to $10$ RBs. It  appears that recurrent architecture sounds more useful for global recovery tasks where the observation matrix entangles the image pixels. Perhaps by going to $k$-space ResNet can better learn the proximals to invert the map. This is deferred to future research. 



\vspace{-2mm}

\subsubsection{Interpretation of generator outputs}
\label{subsubsec:gen_interpretation}
Fig.~\ref{fig:fig_4outputs_5RBs_batc12_superresolution} depicts the output of different generator copies when training a network architecture with $4$ independent copies, each composed of $5$ RBs. RGB images are shown in gray scale. It is seen that different copies focus on features at different levels of abstraction associated with different frequency components. The first block tries to retrieve major (low-frequency) structural features at the expense of introducing a large amount of high-frequency noise, which is then washed away by the next copy. The third copy then adds up high-frequency components to improve the sharpness, which introduces some noise that is again alleviated by the fourth copy to retrieve the output. The overall process tend to alternate between sharpening and smoothing.




\section{Conclusions and closing remarks}
\label{sec:conc}
This paper caters a novel proximal learning framework for automated recovery of images from compressed linear measurements. Unrolling the proximal gradient iterations, a recurrent/cascade architecture is devised that alternates between proximal projection and data fidelity. ResNets are adopted to model the proximals, and a mixture of pixel-wise and perceptual costs used for training. Experiments are examined to assess various network wirings in reconstructing MR images of pediatric patients, and superresolving face images. Our observations indicate that a recurrent small ResNet can effectively learn the proximal, and significantly improve the quality and complexity of recent deep architectures (single copy) and the conventional CS-MRI. Our preliminary results for single-image suprerresolution however indicate that the recurrent architecture are not that effective compared with the exiting deep schemes.

There are still unanswered questions that are the focus of our current research. They pertain to running more experiments with perceptual costs with a subjective quality-assessment strategy; more extensive experiments for superresoltuion with larger ResNet sizes and possibly training in the $k$-space with more global measurements; and a fair mechanism to compare the inference/training time.


{\small
\bibliographystyle{ieee}
\bibliography{egbib}
}

\end{document}